\documentclass[letterpaper, 10 pt, conference]{ieeeconf}
\IEEEoverridecommandlockouts  
\usepackage{cite}
\usepackage{amsmath,amssymb}
\usepackage{graphicx}
\usepackage{booktabs}
\usepackage{hyperref}
\usepackage{siunitx}
\usepackage{tikz}
\usetikzlibrary{shapes,arrows,positioning}

\begin{document}

\title{\LARGE \bf
ARRC: Advanced Reasoning Robot Control---Knowledge-Driven Autonomous Manipulation Using Retrieval-Augmented Generation
}

% \author{
%   \IEEEauthorblockN{XXXX\thanks{Work done while at Research Lab X. Email: XXXX@example.com}}
%   \IEEEauthorblockA{Department of Robotics\\ University / Organization}
% }
\author{Eugene Vorobiov, Ammar Jaleel Mahmood, Salim Rezvani, and Robin Chhabra% <-this % stops a space
% \thanks{*Tmahis work was not supported by any organization}% <-this % stops a space
\thanks{Authors are with the Department of Mechanical, Industrial and Mechatronics Engineering, Toronto Metropolitan University, Toronto, Canada
        {\tt\small evorobiov@torontomu.ca}}%
\thanks{
        {\tt\small ammar.j.mahmood@torontomu.ca}}%
\thanks{
        {\tt\small salim.rezvani@torontomu.ca  }}%
        \thanks{
        {\tt\small robin.chhabra@torontomu.ca  }}%
}

\maketitle
\thispagestyle{empty}
\pagestyle{empty}

\begin{abstract}
We present ARRC (advanced reasoning robot control), a practical system that connects natural language instructions to safe, local robotic control by combining Retrieval-Augmented Generation (RAG) with RGB--D perception and guarded execution on an affordable robot arm. The system indexes curated robot knowledge (movement patterns, task templates, and safety heuristics) in a vector database, retrieves task-relevant context for each instruction, and conditions a large language model (LLM) to synthesize JSON-structured action plans. These plans are executed on a UFactory xArm 850 fitted with a Dynamixel-driven parallel gripper and an Intel RealSense D435 camera. Perception uses AprilTags detections fused with depth to produce object-centric metric poses; execution is enforced via a set of software safety gates (workspace bounds, speed/force caps, timeouts, and bounded retries). We describe the architecture, knowledge design, integration choices, and a reproducible evaluation protocol for tabletop scan/approach/pick--place tasks. Experimental results are reported to demonstrate efficacy of the proposed approach. Our design shows that RAG-based planning can substantially improve plan validity and adaptability while keeping perception and low-level control local to the robot.
\end{abstract}

\section{Introduction}
Robotic manipulation is inherently dual in nature: it requires both high-level reasoning about objects and tasks, and precise low-level execution under physical constraints. Classical robotic systems are strong in motion planning and safety but brittle when adapting to new tasks or linguistic instructions. In contrast, vision-language-action (VLA) models such as RT-1 \cite{BrohanRSS23}, PaLM-E \cite{Driess2023PaLME}, and RT-2 \cite{Chen2023RT2} excel in generalization and zero-shot task understanding, yet they frequently propose unsafe or infeasible plans due to limited embodiment grounding and lack of procedural rule enforcement.

Several works have sought to address these limitations. {SayCan}~\cite{ahn2022can} integrates affordance estimators with language models to filter infeasible actions, while {CLIPort}~\cite{Shridhar2021CLIPort} fuses semantic embeddings with transport operators for spatially grounded pick-and-place. {VIMA}~\cite{Jiang2022VIMA} leverages multimodal prompts to generalize across manipulation tasks with few demonstrations. Although these approaches improve task grounding, they remain constrained by their training distributions and cannot readily incorporate new procedural or safety knowledge without retraining.

Retrieval-augmented generation (RAG), originally proposed for natural language processing (NLP) \cite{Lewis2020RAG}, enables dynamic grounding of model outputs in external, updatable knowledge sources. Recently, robotics researchers have begun exploring retrieval-based compliance frameworks. SayComply \cite{11128684}, for instance, uses retrieval to enforce adherence to operational manuals in field robotics, demonstrating improved compliance and correctness. However, such approaches often stop at high-level compliance and do not integrate low-level metric grounding, perception-driven plan validation, or safety gating for execution.

In this paper, we bridge these gaps by introducing a RAG-enabled robotic manipulation pipeline--called ARRC (advanced reasoning robot control)--that unifies perception, retrieval, and safe plan execution. We deploy this system on a UFactory xArm 850 equipped with a RealSense D435 and a Dynamixel gripper, integrating retrieval of robot-centric safety heuristics and procedural templates at inference time. This design offers both adaptability and reliability, enabling the injection of new task knowledge or safety rules without retraining. Our contributions are as follows:
\begin{itemize}
    \item We develop a hybrid VLA architecture integrating RAG for dynamic injection of procedural and safety knowledge.
    \item We demonstrate a real-world implementation with RGB-D perception, JSON-structured plan generation, and strict execution safety gates.
    \item We propose a reproducible ablation protocol to quantify contributions of retrieval, vision gating, and safety checks.
\end{itemize}

\section{Related Work}

\subsection{End-to-End and Vision-Language-Action Models}
End-to-end VLA models have made significant strides in enabling robots to map visual observations and natural language instructions directly to motor actions. RT-1 (“Robotics Transformer”) \cite{BrohanRSS23} trains a large-scale transformer on over 130,000 robot episodes across diverse tasks, demonstrating strong generalization to new objects and unseen task variations. RT-2 \cite{Chen2023RT2} and PaLM-E \cite{Driess2023PaLME} extend this paradigm by integrating large multimodal models with robotic control, achieving impressive zero-shot capabilities. Other notable efforts include VIMA \cite{Jiang2022VIMA} and SayCan \cite{ahn2022can}, which integrate language reasoning with affordance estimation or reward modeling. LM-Nav \cite{Shah2023LMNav} further highlights the potential of language-conditioned navigation by combining pretrained vision-language-action models with planning modules for long-horizon mobility. Inner Monologue \cite{Huang2023InnerMonologue} explores reasoning and correction through explicit language feedback during execution, enabling agents to adjust plans dynamically. Recent work on using GPT-4 for robotics \cite{Wu2024GPT4Robotics} demonstrates how general-purpose LLMs can guide robotic task planning when provided with appropriate grounding and safety filtering. While these methods excel at leveraging large datasets to generalize across familiar tasks, they remain limited when new safety rules, procedural knowledge, or task constraints must be introduced post-training, often requiring expensive retraining or fine-tuning. Additionally, many end-to-end pipelines do not explicitly enforce low-level safety constraints, which can be critical in real-world robotic deployments.

\subsection{Affordance, Skill Library, and Hybrid Methods}
Hybrid approaches attempt to combine learned policies with structured skill libraries or affordances. CLIPort \cite{Shridhar2021CLIPort} fuses visual-semantic embeddings with transport operators for spatially grounded pick-and-place, enabling more sample-efficient manipulation in tabletop settings. VIMA \cite{Jiang2022VIMA} leverages few-shot multimodal prompts to extend policy generalization across manipulation tasks. SayCan \cite{ahn2022can} couples high-level LLM plans with affordance and value estimators to filter infeasible actions, producing safer and more practical execution sequences. Neural task programming \cite{8460689} and skill-chaining approaches \cite{8460756} encode reusable motion primitives for modular policy composition. More recent methods, such as Huang \textit{et al.} \cite{Corti2023LMZeroShot}, show that LLMs can act as zero-shot planners for robotic tasks, though they require additional grounding for feasibility. LLM-BT \cite{Khandelwal2023LLMBT} embeds structured behavior trees with large language models, improving task decomposition and policy reusability. While these approaches improve feasibility and generalization within known domains, they often rely on fixed libraries or pre-defined heuristics, limiting adaptability when new tasks, safety rules, or procedural knowledge need to be integrated dynamically.

\subsection{Retrieval-Based and Compliance-Grounded Planning}
Retrieval-augmented generation (RAG) \cite{Lewis2020RAG} in NLP improves grounding by combining a language model with an external knowledge corpus. Robotics researchers have begun adapting these ideas for compliant, knowledge-grounded planning. SayComply \cite{11128684} retrieves compliance rules and manuals to constrain LLM outputs, improving task correctness and adherence to operational procedures. Similarly, behavior tree embeddings \cite{Cao2022BTEmbeddings} capture structured task-level knowledge to support retrieval, task comparison, and indexing, but they do not integrate real-time perception or low-level control for safety-aware execution. Other related works include structured affordance retrieval \cite{gutierrez2022thinking}, multi-step plan synthesis with semantic grounding \cite{shridhar2023perceiver}, and Inner Monologue-style online corrections \cite{Huang2023InnerMonologue}, which improve execution reliability but often do not incorporate validated real-robot execution or dynamic safety checks. Our approach extends these paradigms by combining retrieval-augmented LLM planning with local perception, validated execution, and low-level safety enforcement, allowing for flexible adaptation to new tasks, objects, and procedural constraints in real-world robotic environments.

\subsection{Gaps and Motivation for RAG-Driven Execution}
From the survey, we identify several limitations in existing approaches. End-to-end VLA models exhibit strong generalization but are prohibitively expensive to update when introducing new tasks or safety constraints. Hybrid and affordance-based methods alleviate infeasibility, yet they remain restricted by fixed skills and affordance libraries. Retrieval-based compliance systems such as SayComply enforce high-level rules, but they rarely integrate metric perception, plan validation, or execution safety. Similarly, task-structure embeddings (e.g., BT embeddings) enable efficient knowledge indexing but often lack actionable primitives and safety gating within execution pipelines. To overcome these limitations, we present a new framework that unifies retrieval-augmented generation with robot-centric knowledge, metric perception, strict safety gating, and validated execution on real hardware—delivering both the flexibility of language-driven systems and the reliability required for safe robotic autonomy.

\section{System Overview}
%Figure~\ref{fig:system} illustrates the overall architecture, which is composed of three interacting subsystems, reviewed below.

\subsection{Perception}
AprilTags, combined with depth data from the Intel RealSense D435, provide marker-based detections that are fused to recover metric 3D poses in the robot frame. With sub-pixel accuracy, AprilTags enable robust and precise pose estimation, making them well-suited for manipulation tasks~\cite{Olson2011AprilTags}.

  \subsection{Retrieval \& Planning} We construct a curated robotics knowledge base comprising movement primitives, templates, safety heuristics, short demonstration transcripts, and parameterized affordances. This knowledge base is embedded using a SentenceTransformers model and indexed in ChromaDB (or a FAISS index). At inference time, the retrieval module selects the top-$k$ relevant context snippets, which are concatenated with the current observation summary and provided to an LLM (e.g., Gemini, PaLM-E-style models). The LLM then generates a structured JSON plan, enabling downstream execution~\cite{Reimers2019SentenceBERT}.
  \subsection{Execution} The JSON plan—represented as a sequence of named actions with bounded parameters—is first validated by a plan checker, synchronized with the latest object observations, and then executed through the XArm Python SDK. Execution is safeguarded by software safety gates, including workspace and joint limits, capped Cartesian speeds and accelerations, gripper torque/time gating, per-step timeouts, and bounded retries. Low-level controllers and perception remain local to the robot, while the LLM planner can be configured to run either locally (on-device) or via cloud APIs.

\section{Proposed Method}
\label{sec:method}
In this section, we detail each system component and our design choices.

\subsection{Perception: AprilTags \& Depth Fusion}
The perception pipeline employs AprilTag detection (TagStandard41h12 family) to generate marker-based detections with unique object IDs. For each tag, we compute a robust depth estimate by taking the median of depth pixels within the detected tag region. This depth is back-projected through the calibrated camera intrinsics to obtain a 3D point in camera coordinates, which is subsequently transformed into the robot base frame via the extrinsic calibration ${}^{\text{base}}\!T_{\text{cam}}$. Each detected object is published as a compact observation message containing \{\texttt{tag\_id}, \texttt{bbox}, \texttt{position\_xyz}, \texttt{confidence}\}. A lightweight temporal filter smooths frame-to-frame jitter, reducing flicker and improving stability of the reported detections.

\subsection{Knowledge Base and Retrieval}
The knowledge base is designed as a structured collection of short textual entries (50–400 tokens) that capture reusable domain knowledge for robotic manipulation. These entries span movement primitives (e.g., hover approaches, grasp offsets, retreat motions), parameterized task templates (such as \textsc{scan area} $\rightarrow$ \textsc{approach} $\rightarrow$ \textsc{grasp} $\rightarrow$ \textsc{retreat}), safety heuristics (safe hover heights, per-object velocity or force limits, and recovery strategies), as well as affordance notes derived from short demonstrations (e.g., “grasp handle from the side and lift with slow ascent”).

Each document is embedded using a SentenceTransformers model from the Sentence-BERT family and stored in a vector index. In our implementation we employ ChromaDB, though FAISS or similar vector stores can serve as alternatives. At inference time, natural language instructions—optionally fused with compact visual summaries—are embedded and used to query the index. The retrieval step returns the top-$k$ semantically relevant documents, which are then passed forward to condition the planning module \cite{ChromaDB2023}.

\subsection{RAG Prompting and JSON Plan Synthesis}
The retrieved knowledge snippets are concatenated into a structured prompt that also incorporates a compact environment summary (object classes with recent positions), explicit system constraints (workspace limits, maximum speeds), and a JSON plan schema with few-shot exemplars. This prompt is submitted to a LLM, which in our prototype was accessed via a cloud API but can be replaced with any compatible model. The LLM generates a JSON-structured plan consisting of a goal description, optional reasoning metadata, and a sequence of action steps with bounded parameters (e.g., \texttt{APPROACH\_OBJECT} \{label:``bottle", hover\_mm:30, timeout\_sec:8\}). By grounding generation in retrieval-augmented context, the system improves planning specificity and adaptability without retraining. The resulting plan is then parsed and validated by the plan checker before execution.

\subsection{Plan Representation, Validation, and Safety}
Plans follow a restricted JSON schema:
\begin{verbatim}
{
  "goal": "place bottle on tray",
  "steps": [
    {"action":"SCAN_AREA", 
     "params":{...}},
    {"action":"APPROACH_OBJECT", 
     "params":{"label":"bottle",
               "hover_mm":40}},
    {"action":"MOVE_TO_POSE", 
     "params":{"xyz_mm":[...],
               "rpy_deg":[...]}}
  ]
}
\end{verbatim}
Before execution, each action step is validated against system constraints, including parameter bounds such as speeds and positions, and overall plan length. Steps that require current environmental information are synchronized with the latest perception data, and high-risk operations may optionally include human-in-the-loop confirmation. During execution, the executor enforces runtime safety through mechanisms such as per-step timeouts, gripper aborts triggered by load or duration limits, and emergency retreat in response to repeated failures.

\subsection{Gripper, Actuation, and Communication}
We employ a parallel-jaw gripper actuated by Dynamixel motors (Protocol 2.0), providing precise and repeatable grasping for manipulation tasks. The software supports two configurations: a generic MX-series Dynamixel setup that allows direct control over position, speed, and torque, and an optional simplified driver tailored for the XL330-M288 model. Gripper parameters—including servo IDs, communication baudrate, open and closed positions, speed and torque limits, and load thresholds—are defined in \texttt{config/gripper/gripper\_config.yaml}, allowing flexible adjustment for different tasks and objects. The controller enforces safe operational bounds on all commands and integrates load- and time-based gating to automatically detect grasp success or failure. This ensures that grasping actions remain robust under varying object geometries, weights, and environmental conditions, while preventing damage to the hardware.

\subsection{Executor API and Safety Gates}
The executor exposes a compact set of high-level actions that the JSON planner can invoke, including
\texttt{SCAN\_AREA}, \texttt{APPROACH\_OBJECT}, \texttt{MOVE\_TO\_POSE},
\texttt{OPEN\_GRIPPER}, \texttt{CLOSE\_GRIPPER}, \texttt{RETREAT\_Z}, etc.
Each high-level action is mapped to a sequence of atomic XArm SDK commands with pre-validated parameters. Runtime safety is enforced through multiple mechanisms: workspace boundaries and maximum single-move distances are respected, Cartesian and joint velocities and accelerations are capped, and gripper operations are monitored via load- and time-based thresholds with emergency open procedures triggered on anomalies. Each step also supports bounded retries, and overall plan execution is subject to a configurable timeout.

Critically, the executor operates as the final stage of our end-to-end pipeline, translating LLM-generated JSON plans—conditioned on retrieval-augmented knowledge and real-time perception—into safe, executable robot motions. All safety parameters, including limits, thresholds, and retry counts, are fully configurable, allowing the system to adapt to different manipulators, end-effectors, and task environments. This integration ensures that high-level reasoning and planning performed by the LLM results in robust, reliable, and safe manipulation in real-world scenarios.

\section{Algorithmic Processes}
\label{sec:algorithm}

\begin{figure*}[t]
  \centering
  \includegraphics[width=13cm]{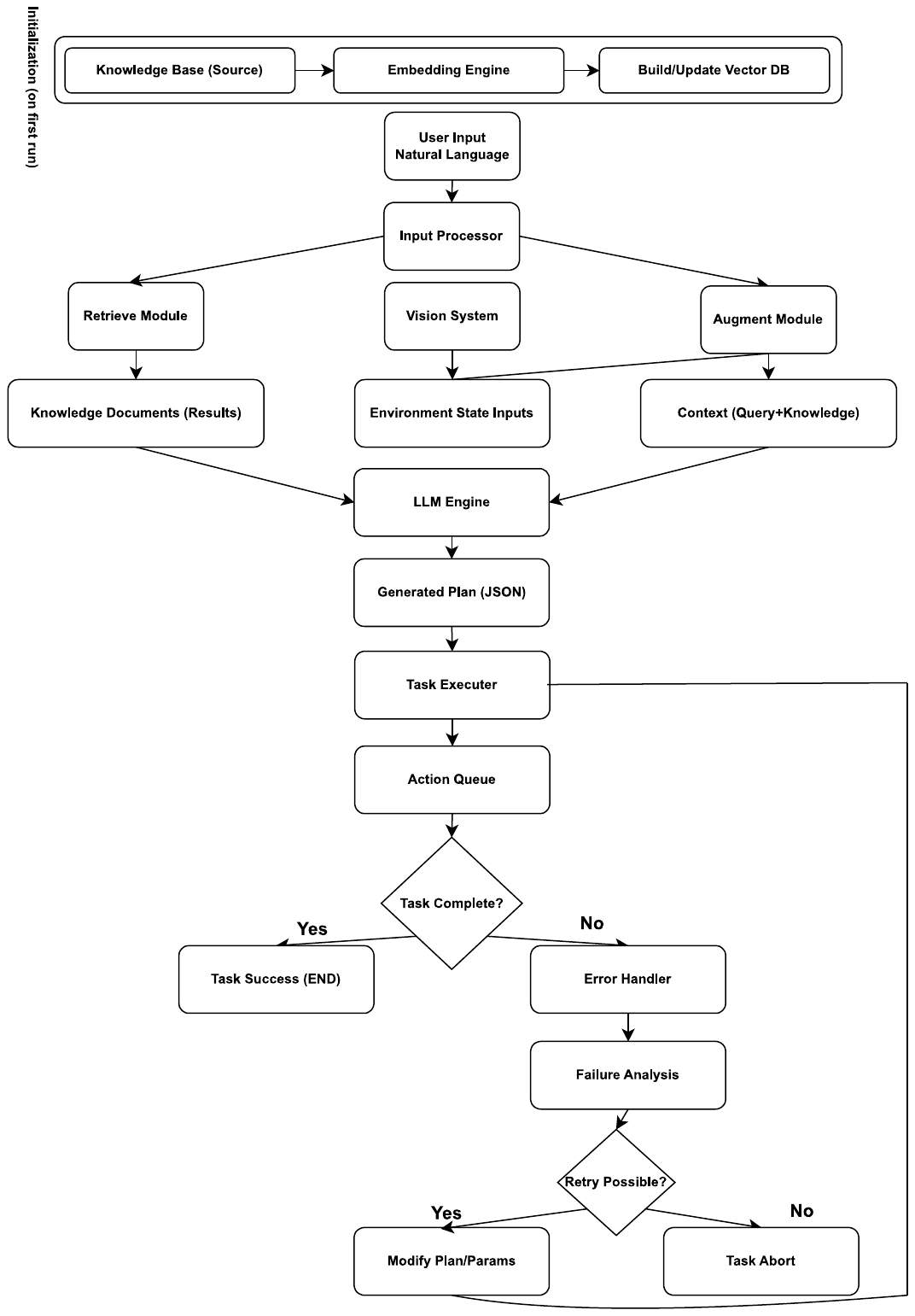}
  \caption{High-level architecture. Perception produces object-centric observations; the RAG planner retrieves task knowledge and synthesizes a JSON plan; the executor validates and executes actions through the XArm SDK with safety gates.}
  \label{fig:system}
\end{figure*}

This section details the core algorithmic components that enable the system's reasoning and execution capabilities. All the steps are described in the Figure.\ref{fig:system}.

\subsection{RAG-Based Planning Algorithm}
The planning process follows a structured retrieval-augmented generation pipeline.

\begin{enumerate}
\item \textbf{Query Processing}: a natural-language instruction is embedded with SentenceTransformers to produce a query vector $q \in \mathbb{R}^d$.
\item \textbf{Knowledge Retrieval}: The query is compared against the knowledge-base embeddings $\{k_i\}_{i=1}^N$ using cosine similarity:
\begin{equation}
\text{sim}(q, k_i) = \frac{q^\top k_i}{\|q\| \|k_i\|}
\end{equation}
The top-$m$ most similar knowledge entries are retrieved and stored in the vector $\mathcal{R}$.
\item \textbf{Context Assembly}: Retrieved knowledge entries are concatenated with a symbolic summary of the current environment state $s_t$ (e.g., object classes and poses) to form the full planning context $C = [\mathcal{R}^\top, s_t^\top]^\top$.
\item \textbf{Plan Generation}: The context $C$ is passed to an LLM, which produces a JSON-structured plan $\pi = \{g, A\}$, where $g$ is the high-level goal and $A = \{a_1, \dots, a_l\}$ is a sequence of actions (action queue). Each action is represented as a tuple of an action label and bounded parameters (e.g., \texttt{APPROACH\_OBJECT \{label: "bottle", hover\_mm: 30, timeout\_sec: 8\}}). 
\end{enumerate}

\subsection{Hierarchical Scanning Algorithm}
The manipulator system employs a two-phase scanning routine for robust object detection. In the first phase, a \textbf{horizontal scan} is performed across the workspace at a fixed height, where the perception module inspects each sampled position for potential targets. If no objects are detected, the system transitions to a fallback \textbf{arc scan}, which uses three predetermined joint-space configurations (\emph{LEFT}, \emph{CENTER}, and \emph{RIGHT}). These configurations are hardcoded to ensure safe, repeatable coverage of the workspace and avoid kinematic singularities, providing a deterministic alternative to Cartesian waypoint-based exploration.

\subsection{Coordinate Transformation Pipeline}
Object coordinates obtained from AprilTags are first expressed in the eye-in-hand camera frame and then transformed into the robot base frame using calibrated transformations. These base-frame coordinates are subsequently used for inverse kinematics computations during manipulation. The transformed coordinates are stored in memory and retrieved on demand, depending on the specific objects referenced in the execution plan generated by the LLM.

We obtain the AprilTag position in the camera frame as $P_a^{\text{cam}}\in\mathbb{R}^3$, whenever the tag is in the field of view of the camera. The corresponding object position in the manipulator base frame $P_a^{\text{base}}\in\mathbb{R}^3$ is then computed via calibrated homogeneous transformations. 
\begin{align}
    \begin{bmatrix}
        P_a^{\text{base}}\\ 1
    \end{bmatrix} = H_{\text{TCP}}^{\text{base}}(\theta)H_{\text{cam}}^{\text{TCP}}\begin{bmatrix}
        P_a^{\text{cam}}\\1
    \end{bmatrix},
\end{align}
where $H_{\text{TCP}}^{\text{base}}$ is the $4\times 4$ forward kinematics transformation from the Tool Center Point (TCP) to the base frame, parameterized by the joint configuration $\theta,$ and $H_{\text{cam}}^{\text{TCP}}$ is the fixed, calibrated extrinsic transformation from the eye-in-hand camera to the TCP. 

Once the transformed position $P_a^{\text{base}}\in\mathbb{R}^3$ is obtained, it is validated against the robot’s workspace bounds identified by the robot's reachable $\mathcal W$. Specifically, the position is checked to ensure it lies within the reachable volume of the manipulator, does not violate joint limits, and respects task-level safety constraints (e.g., avoiding collisions with the table or exceeding height limits). Only poses that pass these checks are forwarded to the inverse kinematics solver and subsequently used in the execution of the planned task.

% \begin{enumerate}
% \item \textbf{Depth Fusion}: To compute robus depth statice, for each detected April Tag transformation $T_i$, we use the following filter:
% \begin{equation}
% \bar T_{i} = \text{median}(\{T_{ij} : (i,j) \in \text{tag\_region}\})
% \end{equation}
% where $d_{ij}$ are depth values within the detected tag bounding box.

% \item \textbf{Camera to Base Transformation}: Apply calibrated extrinsic transformation:
% \begin{equation}
% \begin{bmatrix} x_{\text{base}} \\ y_{\text{base}} \\ z_{\text{base}} \\ 1 \end{bmatrix} = {}^{\text{base}}T_{\text{cam}} \begin{bmatrix} x_{\text{cam}} \\ y_{\text{cam}} \\ z_{\text{cam}} \\ 1 \end{bmatrix}
% \end{equation}
% where ${}^{\text{base}}T_{\text{cam}}$ is the 4×4 transformation matrix from camera to robot base frame.

% \item \textbf{Pose Validation}: Check if transformed pose $(x_{\text{base}}, y_{\text{base}}, z_{\text{base}})$ lies within workspace bounds $\mathcal{W}$ and is reachable by the robot.
% \end{enumerate}

\subsection{Safety-Constrained Execution}
The execution module enforces a set of safety constraints to guarantee reliable operation during plan execution. First, all commanded target positions are validated against the robot’s reachable workspace, ensuring all $P_a\in\mathcal W$.
Motion commands are further regulated through velocity limits: Cartesian translations satisfy $\|\dot P\|\leq v_{\text{max}}$, while joint velocities are constrained by $|\dot\theta_i|\leq\dot\theta_{i,\text{max}}$. Each action step is assigned a maximum allowable duration $t_{\text{max}}$; exceeding this limit triggers an abort and initiates a return-to-safe configuration. Gripper operations are protected through real-time load monitoring, with an emergency release executed if excessive force or unexpected obstruction is detected.

By integrating workspace validation, dynamic constraints, temporal bounds, and force-aware gripper control, the execution framework achieves robust, safe deployment of RAG-generated plans without sacrificing real-time responsiveness or hardware longevity.

\section{Experimental Setup}
\label{sec:exp}

The experimental protocol is designed to provide a reproducible evaluation of the proposed manipulator system. We describe the hardware and software stack, the benchmark tasks and scenes, and the evaluation metrics used. 

\subsection{Hardware and Software}
Experiments are conducted on a UFactory xArm~850 six-degree-of-freedom manipulator controlled over Ethernet through the official Python SDK. The manipulator is equipped with a parallel-jaw gripper driven by Dynamixel actuators, with both MX-series and an optional XL330 configuration supported. Perception is provided by an Intel RealSense D435 RGB--D camera, which is rigidly mounted and extrinsically calibrated with respect to the robot base frame, ensuring accurate recovery of 3D object poses from AprilTag detections. All computation runs on a host workstation running Ubuntu~22.04 and Python~3.10. The perception system is implemented in \texttt{vision\_system/pose\_recorder.py}, high-level planning is handled by \texttt{rag\_system/true\_rag\_planner.py}, and low-level execution is carried out by \texttt{robot\_controller/executor.py} interfacing with the robot through \texttt{robot\_control/main.py}. ROS~2 is optionally used to publish vision messages, although the entire pipeline can run in a standalone lightweight mode. By default, LLM calls are executed via cloud APIs, but the system can be reconfigured to use a local LLM when available.

Table~\ref{Limitation} summarizes the programmatically enforced constraints and boundaries that regulate system resources and ensure stable performance.

\begin{table}[h]
  \centering
  \caption{Limits enforced in software during execution}
  \begin{tabular}{l l}
    \hline
    Parameter & Value \\
    \hline
    Max Cartesian speed & 150\,mm/s \\
    Max joint speed & 30\,deg/s \\
    Max Cartesian acceleration & 1000\,mm/s$^2$ \\
    Workspace $x$ & [150, 650]\,mm \\
    Workspace $y$ & [-300, 300]\,mm \\
    Workspace $z$ & [50, 500]\,mm \\
    Gripper max speed & 300 (SDK units) \\
    Gripper max force & 100 (SDK units) \\
    Safety check interval & 0.1\,s \\
    Max single move distance & 400\,mm \\
    \hline
  \end{tabular}
  \label{Limitation}
\end{table}

\subsection{Tasks and Scenes}
\begin{figure}[t]
  \centering
  \includegraphics[width=0.4\textwidth]{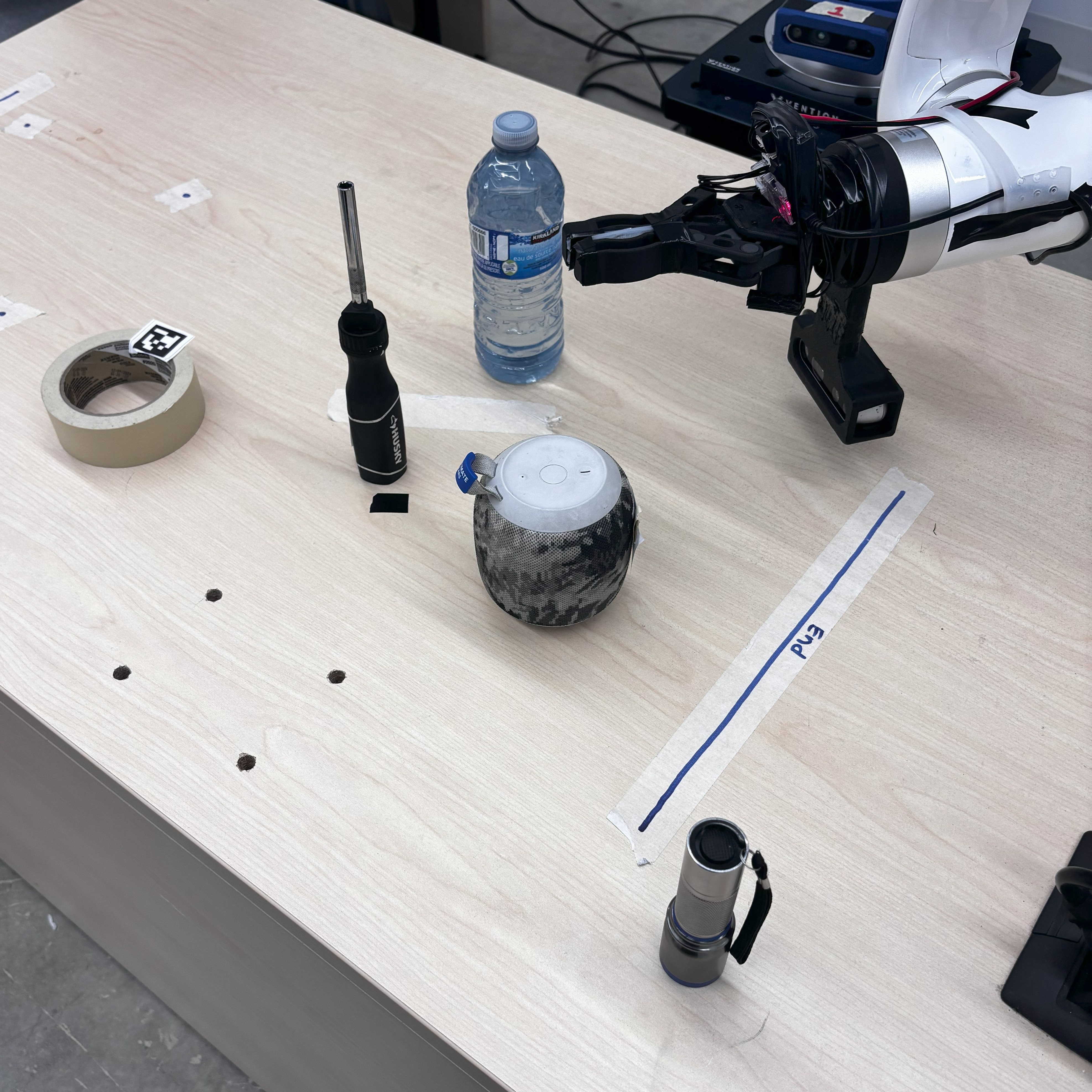} % <-- replace with actual file path
  \caption{A representative scene where the robot is performing manipulation.}
  \label{fig:scene}
\end{figure}

We benchmark the system on three canonical tabletop manipulation tasks that reflect common challenges in domestic and industrial settings. The workspace is populated with a mix of everyday household objects, including bottles, boxes, and hand tools such as screwdrivers, arranged in moderate clutter to ensure that perception and motion planning must handle partial occlusions and close object proximities. All tasks are physically instantiated on a laboratory workbench, with object arrangements randomized across trials while remaining within a bounded workspace region to maintain consistency.  Fig.~\ref{fig:scene} illustrates a representative cluttered scene where the robot performs a manipulation task..

The first task, \textbf{scan}, requires the manipulator to execute a workspace sweep using the mounted depth camera in order to detect and localize visible objects using AprilTags. The outcome is a set of object observations that are published for downstream planning. The second task, \textbf{approach}, evaluates the system’s ability to condition on a class label such as ``bottle'' and to generate a collision-free plan that positions the end-effector at a stable hover pose above the identified object. Finally, the \textbf{pick--place} task tests full manipulation capability: the robot must execute a grasp on the specified object, lift it from the tabletop, transport it to a designated drop location, and release it without premature slippage or collision.  

\begin{figure}[t]
  \centering
  \begin{tikzpicture}[node distance=2cm, auto, scale=0.7]
    \tikzstyle{process} = [rectangle, draw, fill=blue!20, text width=2cm, text centered, rounded corners, minimum height=0.6cm, font=\scriptsize]
    \tikzstyle{decision} = [diamond, draw, fill=yellow!20, text width=1.5cm, text centered, minimum height=0.6cm, font=\scriptsize]
    \tikzstyle{action} = [rectangle, draw, fill=green!20, text width=2cm, text centered, rounded corners, minimum height=0.6cm, font=\scriptsize]
    \tikzstyle{failure} = [rectangle, draw, fill=red!20, text width=2cm, text centered, rounded corners, minimum height=0.6cm, font=\scriptsize]
    \tikzstyle{arrow} = [thick,->,>=stealth]
    
    \node (start) [process] {User: "pick up screwdriver"};
    \node (rag) [process, below of=start] {Retrieve robotics knowledge};
    \node (plan) [action, below of=rag] {Generate scan plan};
    \node (scan) [process, below of=plan] {Horizontal sweep scan};
    \node (check1) [decision, below of=scan] {Screwdriver visible?};
    \node (arc) [process, right of=check1, xshift=2.5cm] {Arc scan: 3 positions};
    \node (check2) [decision, below of=arc] {Found from angle?};
    \node (approach) [action, below of=check2] {Approach with RPY};
    \node (grasp) [action, below of=approach] {Grasp \& lift};
    \node (not_found) [failure, left of=check2, xshift=-2.5cm] {Object not found};
    \node (reprompt) [process, below of=not_found] {Reprompt RAG with failure context};
    \node (new_plan) [action, below of=reprompt] {Generate new plan};
    
    % Main flow arrows
    \draw [arrow] (start) -- (rag);
    \draw [arrow] (rag) -- (plan);
    \draw [arrow] (plan) -- (scan);
    \draw [arrow] (scan) -- (check1);
    
    % Decision arrows with better positioning
    \draw [arrow] (check1) -- node[above] {No} (arc);
    \draw [arrow] (check1) -- node[left] {Yes} (approach);
    
    \draw [arrow] (arc) -- (check2);
    \draw [arrow] (check2) -- node[right] {Yes} (approach);
    \draw [arrow] (check2) -- node[above] {No} (not_found);
    
    % Success path
    \draw [arrow] (approach) -- (grasp);
    
    % Failure recovery path
    \draw [arrow] (not_found) -- (reprompt);
    \draw [arrow] (reprompt) -- (new_plan);
    
    % Feedback loop - much more curved
    \draw [arrow] (new_plan) to[bend left=60] (scan);
  \end{tikzpicture}
  \caption{Adaptive reasoning flow for "pick up the screwdriver" with failure recovery. The system demonstrates intelligent fallback from horizontal to arc scanning, and when objects remain undetected, it reprompts the RAG system with failure context to generate alternative strategies.}
  \label{fig:test_case_flow_compact}
\end{figure}
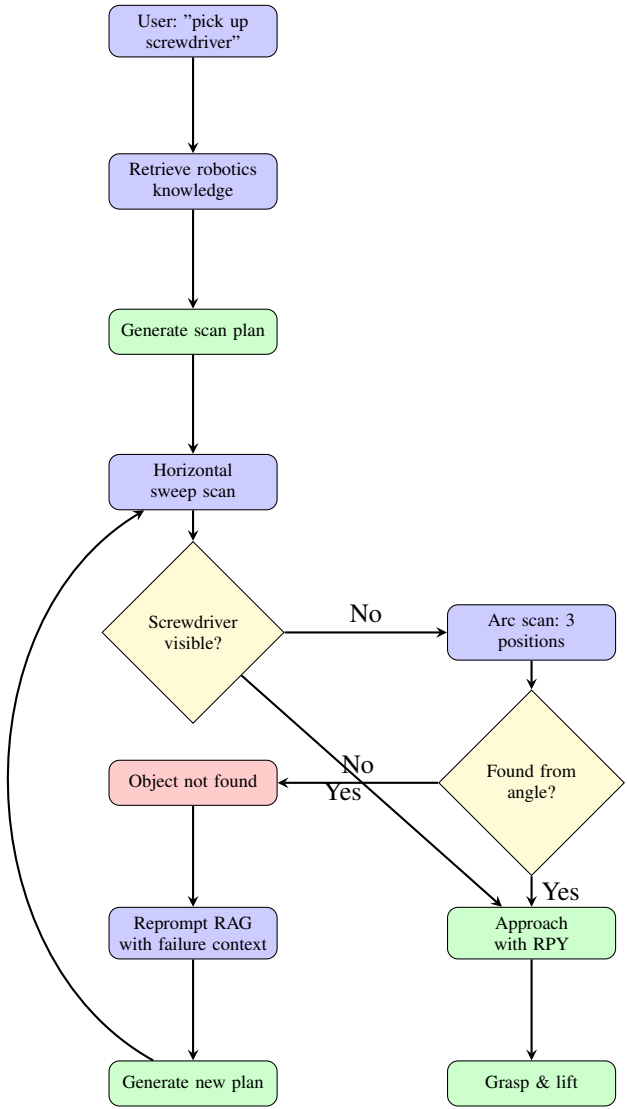

To support reproducibility, scenes are generated from a limited set of objects and placement strategies, with identical layouts repeated across multiple trials for each task. This ensures that performance differences reflect system behavior rather than uncontrolled environmental variation, while still introducing sufficient diversity to challenge perception and planning modules. Fig. \ref{fig:test_case_flow_compact} demonstrates the adaptive reasoning pipeline that was used in the experiments to pickup and place a screwdriver.

% \subsection{Metrics}
% Performance is evaluated across repeated trials using several quantitative measures:  

% \begin{itemize}
%   \item \textbf{Scan Success (\%):} Percentage of trials in which the scan routine detects at least one object above the confidence threshold.  
%   \item \textbf{Approach Success (\%):} Fraction of trials in which the end-effector successfully reaches a valid hover pose within defined positional and orientation tolerances.  
%   \item \textbf{Pick--Place Success (\%):} Proportion of runs in which the target object is grasped, lifted, transported, and released at the designated location without premature drop or collision.  
%   \item \textbf{Plan Validity (\%):} Percentage of generated plans that pass both syntactic and semantic validation before execution.  
%   % \item \textbf{Latency (s):} Wall-clock time, reported both per stage (scanning, planning, approach) and for the complete end-to-end pipeline.  
% \end{itemize}  

% To further analyze system performance, we conduct an ablation study:  

% \begin{itemize}
%   \item \textbf{RAG Off:} Retrieval-augmented generation is disabled, so planning relies only on the natural language input.  
%   \item \textbf{No Vision Gating:} The executor bypasses vision-based updates and instead relies on nominal object poses.  
%   \item \textbf{Safety Caps Off:} Software safety constraints, including workspace limits, velocity caps, and timeouts, are removed to measure the runtime overhead of safeguards.  
% \end{itemize}  

Each condition is repeated for at least ten trials, with results reported as averages and standard deviations.

% \subsection{Baselines}
% \begin{itemize}
%   \item \textbf{Scripted baseline}: hard-coded pick--place with fixed poses (no perception, no RAG).
%   \item \textbf{LLM-only}: a planner that uses LLM without retrieval context (same perception/execution).
%   \item \textbf{Behavior cloning (BC)}: imitation-learned policy where available.
% \end{itemize}

\section{Results}

In the test case, an object was positioned to partially occlude a screwdriver at an angle. The system demonstrated advanced reasoning by: (1) initially performing a horizontal scan that failed to detect the target object due to occlusion, (2) automatically transitioning to an arc scan strategy to view the workspace from multiple angles, (3) successfully detecting the screwdriver from the arc scan position, and (4) executing a successful pick and place operation. This example showcases the system's ability to reason about occlusion and adapt its scanning strategy procedurally.

Representative failures we observed during prototyping include:

  i) missed detections under partial occlusion (recovery: rescan + change point of view),
  
  ii) low-confidence objects causing approach timeouts (recovery: fall back to scripted grasp),
 
  iii) fragile objects requiring gentle close and tactile sensing (not yet instrumented).

Table. \ref{Results} provides a quantitative measure of the system's performance, which directly supports the superiority of the proposed ARRC framework.

\textbf{Plan Validity:} In 8 out of 10 trials, the plan was valid, which suggests the system consistently generates a correct and executable path. The two failures ("No" in trials 3 and 5) align with the "representative failures" mentioned, such as missed detections under partial occlusion, which could lead to a failed plan.

\textbf{Scan:} The "T" in all trials confirms that a target was successfully detected after a scan. This reinforces the example of the system's ability to adapt its scanning strategy to overcome occlusion.

\textbf{Approach (\%):} This is a critical metric for a robotic manipulation task. The "Approach" value represents the accuracy of the robot to place the gripper in the vicinity of the object, which was quantified using the ground truth location of the objects. The high percentages across the board, with an average of 87.1\%, indicate that the system is consistently successful in performing precise pick-and-place operations. This high accuracy is particularly noteworthy given the complexity of the tasks.

\textbf{P\&P:} The "Yes" in 8 out of 10 trials confirms that the pick-and-place operation was successfully executed. The two "No" results correlate with the "Plan Validity" failures, as a failed plan would naturally prevent a successful P\&P execution.

\textbf{Time Frame:} The time frame of the operations is relatively consistent, with an average of 1.225 seconds. This suggests that the system's reasoning and execution are not only accurate but also efficient.

The qualitative observations of system behavior are consistent with the quantitative results reported in Table~\ref{Results}. For example, the screwdriver case under partial occlusion illustrates the system’s ability to perform higher-level reasoning, ultimately yielding successful task execution. This is supported by the high Approach and Pick--Place success rates recorded in the table.  

Conversely, the representative failure modes, such as missed detections under occlusion, account for the two trials in which Plan Validity and Pick--Place were recorded as unsuccessful. The alignment between these qualitative insights and the quantitative measures provides a balanced assessment of both the strengths and current limitations of the proposed system.

\begin{table*}[h]
\centering
\caption{Results of the experimental trials, including plan validity, approach accuracy, and pick-and-place success.}
\begin{tabular}{l l l l l l}
\hline
\textbf{\#} & \textbf{Plan Validity} & \textbf{Scan} & \textbf{Approach Accuracy (\%)} & \textbf{P\&P} & \textbf{Time Frame} \\
 \hline
1 & Yes & T & 86\% & Yes & 1.23 \\
 
2 & Yes & T & 86.80\% & Yes & 1.21 \\
 
3 & No & T & 80.30\% & No & 1.22 \\
 
4 & Yes & T & 88.20\% & Yes & 1.24 \\
 
5 & No & T & 79.30\% & No & 1.21 \\
 
6 & Yes & T & 88.90\% & Yes & 1.23 \\
 
7 & Yes & T & 96.10\% & Yes & 1.23 \\
 
8 & Yes & T & 89.20\% & Yes & 1.23 \\
 
9 & Yes & T & 88.20\% & Yes & 1.22 \\
 
10 & Yes & T & 88.80\% & Yes & 1.23 \\
 \hline
Success Rate. & 80\% & 100\% & 87.1\% & 100\% & 1.225 \\
\hline 
\end{tabular}
  \label{Results}
\end{table*}

\section{Conclusions}

We have presented a practical, retrieval-augmented manipulation system that integrates local RGB--D perception, a vector-indexed robotics knowledge base, and LLM-driven plan synthesis to translate natural language instructions into validated, executable actions on a UFactory xArm 850. Experimental results demonstrate that retrieval-augmented generation improves plan specificity by providing concise, robot-relevant context, enabling more reliable and parameterized plans compared to LLM-only approaches. Furthermore, keeping perception and low-level control local proves effective for real-time execution and safety enforcement, while latency introduced by retrieval and LLM inference can be mitigated through capturing strategies.

Despite these strengths, the current system has some limitations. The knowledge base is curated and static, limiting adaptability to lifelong updates. Tasks requiring tactile feedback or precise torque control, such as dowel insertion or screw driving, remain out of scope. Additionally, the LLM planner may occasionally generate physically infeasible actions, highlighting the necessity for robust plan validation and symbolic feasibility checks. These limitations underscore important avenues for future research.

Future work will focus on scaling the system to more complex manipulation scenarios, including multi-arm coordination, large-scale benchmarking across diverse objects and lighting conditions, and integration of tactile sensing and force-aware motion primitives. On-device acceleration of retrieval and LLM inference is another promising direction to reduce latency and improve autonomy. Overall, our study demonstrates that combining retrieval-augmented language reasoning with local perception and safety-guarded execution provides a practical and reproducible pathway toward robust, adaptable robotic manipulation in real-world environments.

% \section*{Acknowledgments}
% We thank the lab members and engineers who helped prototype the system. (Optional: list funders and institutions.)

\bibliographystyle{IEEEtran}
\bibliography{references}
\end{document}